# The VEP Booster: A Closed-Loop AI System for Visual EEG Biomarker Auto-generation

Junwen Luo[1], Chengyong Jiang [1], Qingyuan Chen [1], Dongqi Han[2], Yansen Wang[2], Biao Yan[1], Dongsheng Li[2], and Jiayi Zhang[1*]

[1]Fudan University
[2]Microsoft Research

July 21, 2024

## Abstract

Effective visual brain-machine interfaces (BMI) is based on reliable and stable EEG biomarkers. However, traditional adaptive filter-based approaches may suffer from individual variations in EEG signals, while deep neural network-based approaches may be hindered by the non-stationarity of EEG signals caused by biomarker attenuation and background oscillations. To address these challenges, we propose the Visual Evoked Potential Booster (VEP Booster), a novel closed-loop AI framework that generates reliable and stable EEG biomarkers under visual stimulation protocols. Our system leverages an image generator to refine stimulus images based on real-time feedback from human EEG signals, generating visual stimuli tailored to the preferences of primary visual cortex (V1) neurons and enabling effective targeting of neurons most responsive to stimuli. We validated our approach by implementing a system and employing steady-state visual evoked potential (SSVEP) visual protocols in five human subjects. Our results show significant enhancements in the reliability and utility of EEG biomarkers for all individuals, with the largest improvement in SSVEP response being 105%, the smallest being 28%, and the average increase being 76.5%. These promising results have implications for both clinical and technological applications.

## 1 Introduction

Electroencephalography (EEG) biomarkers of visual BMI are specific patterns in EEG signals that can reliably capture properties of visual tasks or stimuli. These biomarkers have significant implications for clinical practices, offering new therapeutic and assistive tools [1]. In neuroscience, EEG biomarkers provide avenues for exploring the underlying processes of cognitive status [2, 3] and brain functions [4]. However, identifying reliable and stable EEG biomarkers poses two key challenges: the inherent non-stationarity of EEG signals [5] and the variability of these signals across different individuals [6].

Previously, the mainstream algorithms for discovering visual EEG biomarkers, such as Canonical Correlation Analysis (CCA) [7] and Task-discriminant Component Analysis (TDCA) [8], have achieved significant success in detecting steady-state visual evoked potentials (SSVEPs). In a recent study, the binocular-task-related component analysis (bTRCA) algorithm in visual BMIs achieved impressive information transmission rates [9]. However, the efficacy of these algorithms often diminishes in real-world settings beyond the controlled laboratory environment due to environmental noise [10], variations in the quality of electrode contact, and fluctuations in the user's physical and mental states [11]. These factors contribute to general individual variations in EEG signals and can lead to a decline in performance.

More recently, researchers have explored more advanced techniques, including deep neural networks (DNNs) and pretrained transformer models, to extract EEG biomarkers. DNNs are particularly effective at capturing spatial and temporal dependencies in EEG data, as demonstrated by studies such as [12, 13, 14]. Meanwhile, kwak2017convolutional showed that Convolution neural networks (CNN) can outperform traditional neural networks and state-of-the-art methods like CCA [7] and its variants, achieving classification accuracy of 99.28% in static setting and 94.03% in a relative ambulatory setting. Pre-trained transformer models [15, 16, 17] offer a robust framework for interpreting EEG data by leveraging universal patterns and features that are common across various EEG-based tasks. These models are typically pre-trained on large-scale unsupervised data to enhance their performance on downstream tasks with limited data. However, while DNNs and transformer models can partially address the non-stationary nature of EEG signals, they may not fully capture the complexities in the origins of EEG biomarker. This is because the correlation between some EEG biomarkers and the signals themselves can be indirect, due to attenuation and contamination by background noise. Therefore, a purely signal-based approach may not be sufficient to fully capture the intricacies of EEG biomarkers.

To address the challenges of generating reliable and stable EEG biomarkers under visual stimulation protocols, we present the "VEP Booster". Inspired by previous work [18, 19], this novel closed-loop AI framework combines a pre-trained Generative Adversarial Network (GAN) with an advanced EEG decoder and a latent vector generator. As illustrated in Figure 1, the system is designed to interact dynamically with a human user, providing tailored visual stimuli that adapt to the responses of the user's V1 neurons. By reducing background oscillations in the brain (as shown by the red lines in the right panel of Figure 1) and enhancing the EEG biomarkers of V1 (as indicated by the black lines), this approach significantly improves the quality of the EEG recordings. Moreover, since the system's processes are continuously informed by the user's current brain state and recording setup, the issue of EEG signal variation is effectively mitigated. In summary, our primary contributions are:

• **Algorithm**: We propose a novel closed-loop AI framework which can automatically generate reliable and stable EEG biomarkers under visual stimulation protocols, addressing the major challenges of EEG non-stationarity and individual signal variations.

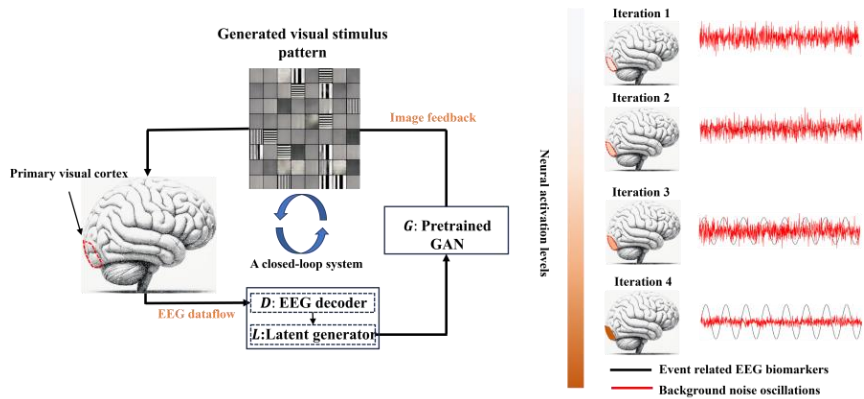

Figure 1: Overview of the VEP Booster from a closed-loop perspective. The system comprises a pre-trained Generative Adversarial Network (GAN), a latent vector generator, and an EEG decoder, as demonstrated on the left-hand side of the figure. As the system interacts with a participant, event-related EEG biomarkers are significantly enhanced, while background noise oscillation is reduced, as demonstrated on the right-hand side of the figure.

- **Implementation**: To demonstrate the feasibility of our proposed framework, we have implemented a comprehensive system that includes a DCGAN, a genetic algorithm-based latent generator, and a robust EEG decoder that uses FFT and SNR techniques.
- **Experiment**: By employing an SSVEP visual protocol, our system is able to elicit EEG biomarker responses from human subjects that is 76.5% more effective than those evoked by natural images, demonstrating the potential of our system in real-word applications.
- **Dataset**: The training dataset and experimental dataset will be open sourced, further contributing to the community for exploring visual brain-machine interfaces.

## 2  The VEP Booster

### 2.1  System Overview

The primary objective of the VEP booster is to generate images that elicit the strongest EEG biomarker responses under a visual stimulus protocol. For a single loop, the system works as follows: 1) a generator generates visual stimulus images; 2) we record the human participant's EEG signals under each visual stimulus protocol with the generated images; 3) an EEG decoder scores the generated images based on the feature values; 4) a latent generator generates the latent inputs of the generator for the next iteration based on the scores from the EEG decoder.

As illustrated in Figure 1, the system comprises of four interconnected sub-systems:

- 1) **Generator**, denoted as $G$; Without loss of generality, we use a pre-trained Generative Adversarial Network (GAN) to implement the generative model $G : Z \rightarrow X$, where $Z$ is the latent space, and $X$ is the space of visual stimuli. $G$ is defined by parameters $\theta_g$, with the optimization objective to minimize the discrepancy between the generated visual stimuli and the dataset visual stimuli.
- 2) **EEG Decoder**, denoted as $D$; Formally, $D$ is the mapping from EEG signals to scores, $D : E \rightarrow \mathbb{R}$, outputting a real-number score indicating the effectiveness of the stimulus. $D$ is defined by parameters $\theta_d$, with the goal to accurately predict the quality and the differences of the visual stimuli corresponding to the given EEG signals.
- 3) **Latent Generator**, denoted as $L$; $L$ be a function that modifies the latent vectors $z$ from the distribution $p(z)$ based on the scores derived from the EEG decoder, formulated by:

$$z_{\text{mutated}} = \arg\max_z \sum_1^n \text{feature}_i(z) + \delta, \quad (1)$$

where $\text{feature}_i(z)$ are functions returning the various feature scores of the EEG signals associated with latent vector $z$, and $\delta$ is a mutation vector where each element $\delta_i$ is drawn from a normal distribution $N(0, \sigma^2)$ with probability $p$, representing the mutation rate. The variance $\sigma^2$ controls the extent of mutation. The number of feature vectors depends on visual protocols.

- 4) **Human Subject**, who actively participates in the closed-loop process. We define an EEG Signal Model $f$, which is the mapping from visual stimuli to EEG signals from the human brain, $f : X \rightarrow E$, where $E$ represents the EEG signal space. This mapping is typically constrained by the neurophysiological processes of the brain and can be highly nonlinear and complex.

The above four sub-systems operate in a closed-loop manner to optimize the following goal:

$$\max_{\theta_g} \mathbb{E}_{z \sim p(z)}[D(f(G(L(z, \alpha); \theta_g); \theta_d))], \quad (2)$$

where $p(z)$ is the prior distribution of the latent space. This goal facilitates the development of a GAN capable of generating visual neuron preferred stimulus from real data by continuously modifying latent vectors based on feedback from the EEG decoder. This forms the basis of a dynamic and iterative learning process that is crucial for maximising the EEG biomarker responses.

## 2.2 System Implementation

To demonstrate the feasibility of our proposed framework, we conduct a case study using a SSVEP stimulation protocol. Our objective is to elicit the most robust EEG biomarker responses under the SSVEP protocol across all five participants. We collected EEG data using a dry wearable device (developed by BrainUp research Lab), which has seven channels and can capture neural activity across a wide visual area.

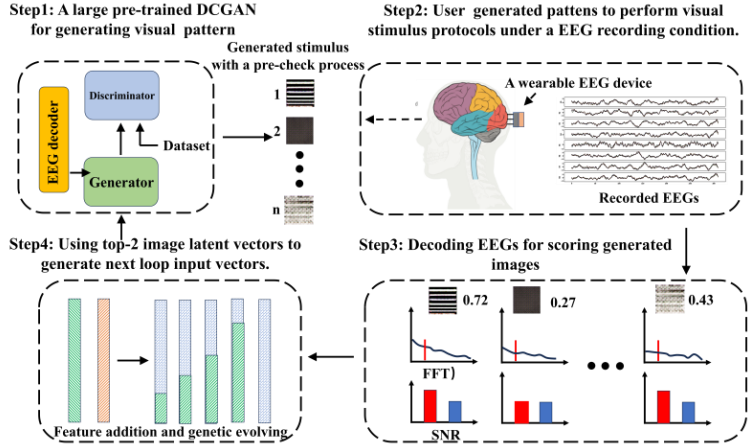

Figure 2: The system implementation. The process involves four main steps: (1) Utilization of a pre-trained Deep Convolutional Generative Adversarial Network (DCGAN) to create visual stimuli; (2) Presentation of these stimuli to subjects during EEG recording sessions, following Steady State Visual Evoked Potential (SSVEP) protocols; (3) Decoding and analysis of the EEG data to evaluate the efficacy of the generated images; (4) Selection and use of the top two performing image latent vectors from the current iteration to refine and generate input vectors for the subsequent iteration.

Our system, depicted in Figure 2, includes a pre-trained Deep Convolutional Generative Adversarial Network (DCGAN) with five layers of specified shapes. To customize the model for our purposes, we create a custom training dataset that featured visual stimuli with variations in brightness, stripe patterns, and checkerboards. Since participants are unable to engage in prolonged closed-loop training, we use a Sinewave-based model to simulate the generation of EEG signals. More details are provided in Appendix A.

The EEG decoder employs a rapid Fast Fourier Transform (FFT) approach, in conjunction with Signal-to-Noise Ratio (SNR) calculations at the target frequency, both as feature extraction techniques. The FFT feature highlights the maximal response within V1 to the target frequency, effectively identifying the most significant neuronal activation triggered by a specific visual stimulus. Conversely, the SNR feature quantitatively assesses the contrast between the amplitude of the target frequency and the energy of the background oscillations. This metric provides insights into the signal's stability, offering an indication of the robustness and clarity of the neuronal response relative to background noise. Based on these two features, a score is calculated for each image using the following equation:

$$\text{Score} = N(\text{FFT}(x(t), f_{\text{target}})) + N(\text{SNR}(x(t), f_{\text{target}})), \quad (3)$$

where $N(x)$ is the normalization function, $x(t)$ represents the EEG data trial in the time domain and $f_{\text{target}}$ is the target frequency.

The latent vector generator algorithm is built based on Equation 1, which combines two latent vectors to produce a single output that delivers optimal performance in FFT and SNR. The best latent vector is then used to generate eight individual offspring vectors, which undergo mutations adhering to a zero-centered Gaussian distribution with a standard deviation ranging from 0 to 80%. Additionally, ten individual offspring vectors are generated using the interpolation process between the best FFT and SNR vectors. This introduces variability and potentially novel feature representations into the generative process.

## 2.3 Training Objective

To ensure that the EEG decoder can identify subtle variations in EEG signals triggered by stimulus images, we defined a loss function to optimize parameters $\theta_d$ and $\theta_l$. This function reflects the discrepancy between the EEG characteristics $f$ induced by the stimulus and the EEG characteristics $f'$ regenerated by the generator $G$. This loss ensures that the refined EEG features are optimized for further processing and integration within the GAN framework.

$$L_{EEG}(f, f') = ||f - f'||^2 \quad (4)$$

where $f'$ is calculated as below:

$$f' = D(G(z_i; \theta_g)). \quad (5)$$

Also, the GAN is designed to explore a wide feature space to enhance the diversity and quality of the biomarkers generated by EEG. The loss function of the GAN is given by:

$$L_{GAN}(y, y_{real}) = \mathbb{E}_{y_{real}}[\log D_c(y_{real}, \theta_d)] + \mathbb{E}_z[\log(1 - D_c(G(z; \theta_g), \theta_d))], \quad (6)$$

where $D_c$ represents the discriminator of the GAN. This loss function ensures that the discriminator can accurately identify between the generated and real stimulus image.

To effectively combine the EEG decoder and GAN-based generation processes, we introduced an integrated optimization objective:

$$\min_{\theta_g, \theta_d} L_{total} = L_{GAN}(y, y_{real}) + \lambda L_{EEG}(f, f'), \quad (7)$$

where $\lambda$ is a weighting factor that balances the importance of the loss of refinement of EEG features against the loss of GAN. This integrated approach ensures that the system optimally refines and generates EEG biomarkers. The details of training process are described in the Appendix A.

# 3 Experiment Results

## 3.1 Experimental Setup

We conducted an experiment involving five human participants (4 male and 1 female). At the first iteration, a trained generator produced 50 images. Through an image precheck process (details are described in Appendix B), the five most diverse images were selected for further testing using a SSVEP protocol at 4 Hz. The reason for selecting 4 Hz is that it is the frequency of the brain's background noise, which aids significantly in evaluating whether the testing system can effectively evoke the target frequency while suppressing the background noise. Each image was presented for a specified duration (125 milliseconds) per trial. Each trial lasted two seconds. After completing 30 trials, participants were allowed a 30-second period to return to resting states. Two best images (decided by a EEG decoder) were then used by the latent generator to produce 20 new images for subsequent iterations in the experiment loop. At each iteration, only five most feature diverse images are selected for an experiment. The total iteration number is eight. Each iteration takes 37 minutes 30 seconds. The experimental picture is shown at Figure 3 (a), depicting a typical experimental configuration, in which a participant outfitted with an EEG device while engaged with a visual stimulus presented on a computer screen. The electrode location is based on 10-20 system: O1, O2,T5, P3, P4, T6 and Pz, which mainly cover V1 areas. The experimental details are described in Appendix C.

## 3.2 SSVEP Response Results

Figure 3 (b) presents an analysis of SSVEP responses, comparing the outputs from the VEP Booster with natural responses. The response (EEG biomarker) is calculated by normalizing two feature values (FFT and SNR) and adding them together (details are shown in Appendix C). The data is calculated by average the results of five human subjects with standard variations.The data illustrates a distinct trend: responses by the VEP Booster (indicated by red triangles) demonstrate a consistent upward trajectory, suggesting an enhancement in the SSVEP responses. Conversely, the naturally observed responses (represented by gray circles) exhibit substantial variability at lower band. At the last iteration loop, the mean SSVEP response generated by the VEP Booster is 76.5% higher than the natural ones, with the improvement ranging from 28% to 105%.

It is noteworthy that there are two turning points in the graph: the initial transient response, which typically exhibits a higher response than subsequent ones due to the adaptation mechanisms of biological systems; and the plateau points, where the SSVEP response stabilizes. A slight decline following the plateau may be attributed to visual fatigue.

Figure 3 (c) illustrates the heat maps of the VEP Booster and natural experimental results of all trials and iterations, respectively. These panels clearly demonstrate that the SSVEP responses increase with each iteration in the VEP Booster simulations, depicted by an escalating intensity in the heat map. In contrast, the heat maps from natural trials show consistently low random values across the iterations. This distinct pattern suggested the capability of the proposed VEP Booster in successfully generating images that preferentially activate biological visual cortex V1, thereby eliciting increasingly strong responses.

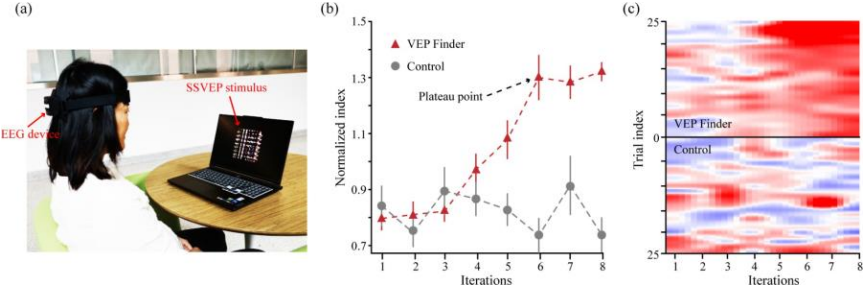

Figure 3: (a) A participant outfitted with an EEG device, actively engaging with a visual stimulus displayed on a computer screen. (b) A comparative analysis of SSVEP (Steady State Visual Evoked Potentials) responses elicited by VEP Booster-generated images versus natural images. The graph presents average values from multiple trials (5 images × 30 trials each) with standard deviations indicated. (c) Heat maps representing the aggregate SSVEP responses to both VEP Booster-generated and natural images from all trial iterations, illustrating the spatial distribution of neural activity.

### 3.3 SSVEP EEGs Results

Figure 4 illustrates a comprehensive view of EEG data and brain topography over eight iterations for a single participant during a visual stimulus experiment. Each panel displays EEG traces with the black line representing the mean amplitude of all trials within that respective iteration. The red vertical dashed line indicates the offset of the visual stimulus. Accompanying each EEG trace, brain topography maps provide a spatial representation of neural activity, emphasizing regions of significant activation. The data reveals a progressive enhancement in signal quality as indicated by the signal-to-noise ratio (SNR) values across the iterations. Starting from an SNR of 1.51 in the first iteration, there is a clear trend of increasing SNR, reaching up to 2.75 in the eighth iteration. Moreover, the amplitude of the EEG responses also shows a significant increase, from 123 $\mu$V in the first iteration to 503 $\mu$V in the eighth, which supports the SNR findings.

### 3.4 EEG Biomarkers Individual Variability

Table 1 summarizes the results of a 4Hz SSVEP EEG experiment, highlighting the impact of our algorithm after eight rounds of optimization. The data demonstrates a consistent improvement in both the signal strength and stability across all subjects. For instance, Subject 1's amplitude increased from 111.10 to 255.21, and his/her SNR improved from 1.32 to 2.66, with a corresponding amplitude enhancement of 129.71% and an SNR improvement of 101%. Similarly, Subject 4 showed substantial gains, with amplitude rising from 245.33 to 503.77 and SNR from 1.51 to 2.76, translating to improvements of 105.35% in amplitude and 82% in SNR. This consistency in the data highlights the system's robustness against the significant variations typically observed in individual EEG responses, which are known for their susceptibility to substantial intra- and inter-individual variability due to factors such as electrode placement, physiological conditions, and environmental influences.

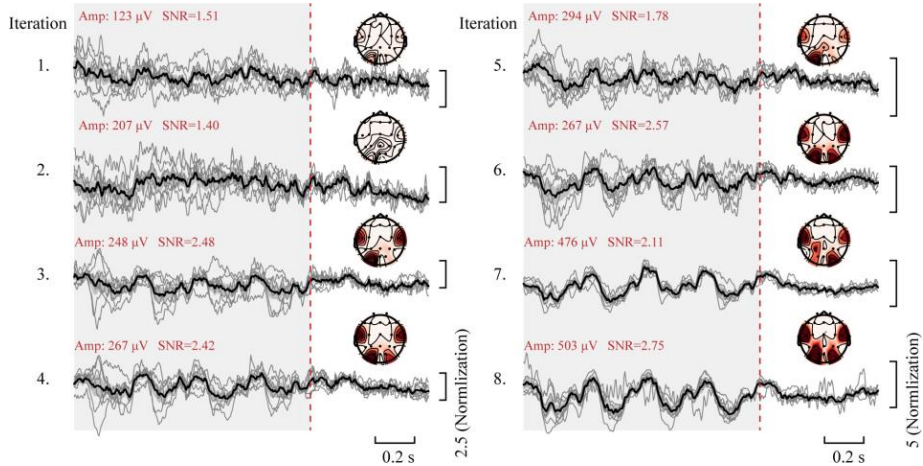

Figure 4: Sequential EEG data and brain topography from eight iterations for a participant. Each panel displays the EEG responses for a single iteration, with the black line representing the average amplitude of all trials within that iteration. The corresponding brain topography maps illustrate the distribution of neural activities. The amplitude (Amp) and signal-to-noise ratio (SNR) for each iteration are noted, demonstrating a progression in response clarity and strength across the sessions.

Table 1: Performance (FFT and SNR) of SSVEP responses at 4Hz.

| Subject | First Iteration (Nature Images) | | Last Iteration | | Performance Improvement | |
| --- | --- | --- | --- | --- | --- | --- |
| | amplitude | SNR | amplitude | SNR | amplitude (%) | SNR (%) |
| 1 | 111.1026 | 1.3218 | 255.2115 | 2.6634 | 129.7080 | 101 |
| 2 | 214.1576 | 2.4627 | 274.1854 | 2.6022 | 28.0298 | 8 |
| 3 | 236.9919 | 1.2181 | 342.7127 | 1.5456 | 44.6095 | 27 |
| 4 | 245.3252 | 1.5122 | 503.7730 | 2.7584 | 105.3490 | 82 |
| 5 | 67.5378 | 1.22 | 156.5173 | 2.209 | 131.74 | 80 |

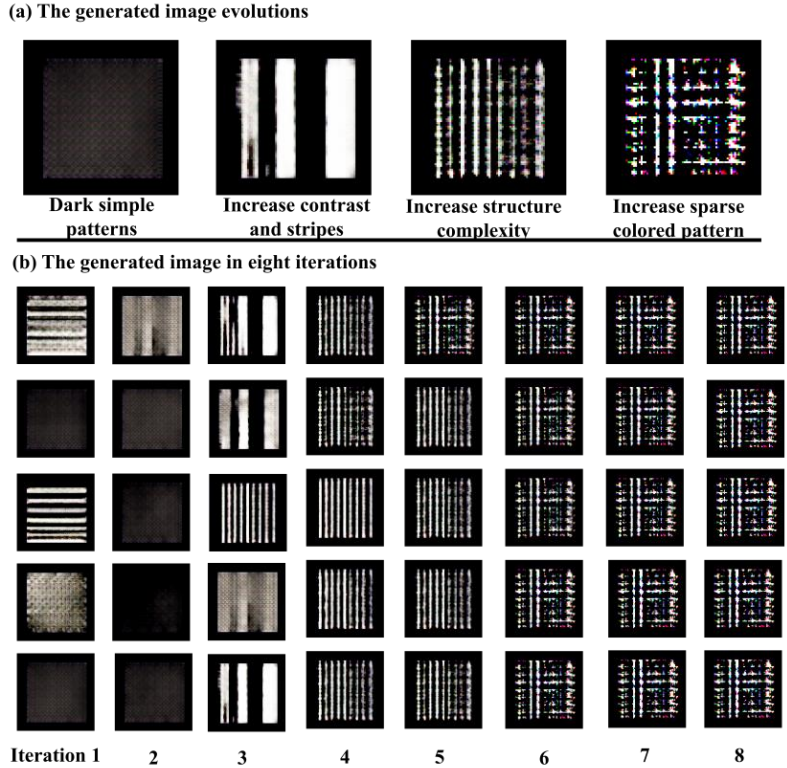

Figure 5: Top: a brief summary of how image evolves. Bottom: progression of synthesized visual stimuli across eight iterations of a participant.

## 3.5 Stimulus Image Evolution

Figure 5 (a) depicts the evolution of generated images across various stages of the experiment. Initially, the images displayed in each column are characterized by simple and fundamental patterns, such as uniform textures, which serve as the baseline for subsequent modifications.

As the experiment progresses through iterations 3 to 4, there is a notable increase in the complexity of the stimuli, with the incorporation of more intricate patterns and enhanced contrast. This evolution is evident as the stimuli develop into configurations that are optimized to elicit robust neural responses.

The experimental sequence reaches its apex in the rightmost images of each column, showcasing stimuli that are not only highly structured and detailed but also feature sparsely colored patterns. These final iterations are posited as the optimal configurations for inducing strong SSVEP responses.

Furthermore, Figure 5 (b) illustrates the outcome of a generator after each iteration loop (after image-checking process, the number is reduced from 20 to 5), displaying a selection of five images that presumably represent the most effective stimuli configurations at each stage.

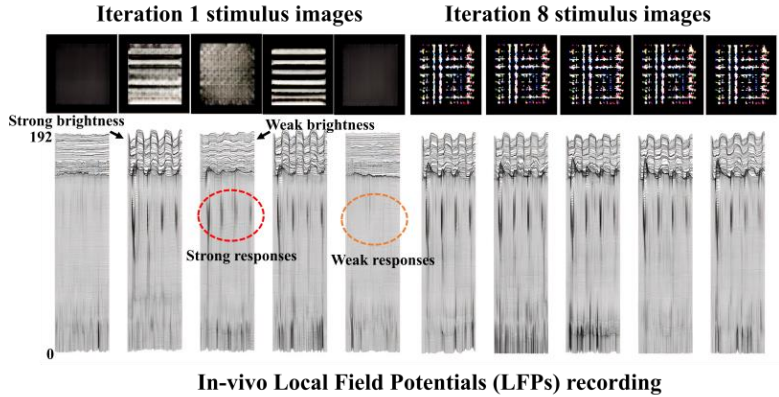

Figure 6: Top row: Visual stimuli from the first and eighth iterations. Bottom row: Corresponding LFP recordings from 192 channels, showing the neural response to each image. The neural responses are categorized into strong and weak, as indicated by red and orange dashed circles respectively, highlighting the variation in neural engagement elicited by different visual stimuli.

### 3.6  *in-Vivo* Animal V1 LFP Recording Verification

To validate the capability of our system to effectively evoke V1 neuron activity, we conducted an *in-Vivo* experiment with mouse, employing invasive Local Field Potential (LFP) recordings under identical SSVEP stimulation protocols as used in non-invasive human studies. The electrodes for these recordings were strategically positioned in the V1 cortical areas. During these experiments, visual stimuli from the first and eighth iterations of our image generation process were presented to the mouse. More details of experiment are described in Appendix D.

As shown in Figure 6, at the eighth iteration, all tested images induced strong LFP responses, closely mirroring the outcomes observed with non-invasive EEG measurements. The x-axis represents a time period spanning 1 second, while the y-axis corresponds to the index of electrode/channel. This alignment underscores the effectiveness of visual stimuli in engaging V1 neuronal activity. Contrastingly, during trials with the first iteration images, only images 2, 3, and 4 showed obvious neural responses in V1 (labelled with red circle). Notably, despite its lower brightness, the third image elicited activities with larger LFP amplitudes, suggesting that structural patterns and the presence of sparse dots within the image also play critical roles in stimulating neuronal responses.

The above characteristics closely align with the image features ultimately synthesised by our system, which integrate optimal brightness, structured patterns, and sparse colored dots. This successful integration confirms that our system can generate images that align closely with the preferences of V1 neurons, enhancing the potential for precise SSVEP-based studies.

## 4 Discussions

### 4.1 Design Principles

Our results have led us to identify several key factors that are critical for the effective operation of the system. First, the optimization of the generator and EEG decoder must be synchronized to ensure that the decoder can detect subtle differences in generated images while the generator explores a rich feature space that reflects the decoder results. This enables fine-tuning of both components, enhancing system performance and ensuring high accuracy in image feature detection and response interpretation. Second, it is essential to balance the GAN loss and the EEG feature loss to prevent either aspect from dominating the learning process, which could result in underfitting one side of the model. The behavior of the GAN loss and the EEG feature loss may differ significantly during training, and monitoring these losses separately can provide insights into model performance and stability. The implementation of the EEG feature loss can be approached in several ways, such as utilizing theoretical models, engaging real users in a feedback loop, or collecting image-EEG data to train a pre-trained model. The choice of method depends on the system requirements and optimization objectives.

### 4.2 Towards a Fine Neural Circuit Activation

In theory, the VEP booster has the potential to activate neural circuits with higher resolution, targeting smaller, more specific neural pathways rather than larger regions such as the V1 cortex. To achieve precise activation of these fine-grained neural features, several considerations are crucial. First, the generative adversarial network must have a sufficiently expansive exploration space coupled with robust feature extraction and integration capabilities [20, 21, 22]. This ensures that the generated stimuli are diverse and detailed enough to selectively activate specific neural circuits. Second, the EEG decoder should be designed to extract a broad range of features from the EEG signals, including capturing dynamic neural patterns [23, 24] and various nonlinear features [25] that are indicative of subtle neural activities. By promoting neuro-plasticity and accelerating rehabilitation processes [26], this technique has the potential to enhance the efficacy of treatments aiming at restoring or improving visual functions [27].

# 5 Conclusion

In this paper, we introduced the "VEP Booster", a novel closed-loop AI framework designed to optimize reliable and stable EEG biomarkers. Utilizing a Steady-State Visual Evoked Potential (SSVEP) protocol, our findings demonstrate that this system can robustly evoke individual biomarkers compared to natural states for all individuals, with the largest increase observed at 105%, the smallest at 28%, and an average improvement of 76.5%, posing the potential of our system in clinical and technological applications.

Regarding **boarder impact**, the capability of the "VEP Booster" to dynamically adapt visual stimuli in response to real-time EEG feedback underscores its significant potential for personalized medicine. This is especially relevant in the context of treating neurological disorders, where conventional approaches that do not account for individual variability may be inadequate. This system's performance highlights its utility in providing tailored therapeutic interventions, promising a shift towards more personalized and effective treatment strategies in neurological care. Also, the development of such AI-driven technologies also stimulates interdisciplinary collaboration between engineers, clinicians, neuroscientists and ethicists, fostering a holistic approach to technological advancement that is ethically sound and socially responsible.

# A Implementation Details

## A.1 Datasets

The dataset was programmatically synthesized using a Python script and consists of images standardized to a resolution of 1024x1024 pixels against a black background. This dataset is categorized into three distinct groups, with each group containing approximately 60 images. The first category encapsulates a gradational luminance shift from black to white. The second category comprises images with uniformly spaced black and white striped patterns. The third category features images with a progressively dense checkerboard grid.

## A.2 DCGAN structure

The discriminator is constructed as a convolutional neural network, specifically designed to process images with dimensions $64 \times 64 \times 3$ (width × height × channels). It incorporates four strided convolutional layers, which sequentially reduce the resolution of the input image. The layers are configured as follows:

- **Layer 1:** Convolutional layer, accepting an input of $64 \times 64 \times 3$, producing an output of $32 \times 32 \times 64$.

- **Layer 2:** Strided convolutional layer, outputting $16 \times 16 \times 128$.

- **Layer 3:** Strided convolutional layer, further reducing to $8 \times 8 \times 256$.

- **Layer 4:** Strided convolutional layer, with a final output of $4 \times 4 \times 512$.

Each convolutional layer, except the first, is followed by batch normalization and a LeakyReLU activation function to improve training stability and introduce non-linearity. The output of the final layer is passed through a sigmoid activation function, yielding a single scalar value. This scalar represents the probability that the input image is classified as real.

The generator in the Deep Convolutional Generative Adversarial Network (DCGAN) framework utilizes a latent vector of length 100, sampled from a standard normal distribution. This vector is processed through a series of transposed convolutional layers to construct an image:

- **Layer 1:** Transposed convolutional layer, converts the latent vector into a $4 \times 4 \times 1024$ feature map.

- **Layer 2:** Transposed convolutional layer, upscales to $8 \times 8 \times 512$.

- **Layer 3:** Transposed convolutional layer, further enlarges to $16 \times 16 \times 256$.

- **Layer 4:** Transposed convolutional layer, increases resolution to $32 \times 32 \times 128$.

- **Final Layer:** Transposed convolutional layer, producing a full-resolution image of $64 \times 64 \times 3$.

Batch normalization and ReLU activations are used in each layer, except in the final layer, where a tanh activation function is used to normalize the image pixels between -1 and 1. This architecture allows the generator to transform a simple distribution into complex data structures that mimic the training images. All experiments were conducted on a computer running Linux Ubuntu 22.04 system, Python version 3.8 and pyTorch 1.10 . The hardware components are as follows: CPU: Intel Xeon 6230 2.1GHz MEM: DDR4 1 TB DISK: 8 TB GPU: Quadro RTX 6000, 24 GB.

### A.3 Closed-loop training

Due to the participants' inability to engage in prolonged closed-loop training, we utilized a model based on Sinewave to simulate the generation of electroencephalographic (EEG) signals under Steady State Visual Evoked Potentials (SSVEP) conditions. This model participated in the closed-loop training process. Within the model, two parameters were set to generate sine waves corresponding to variations in image luminance and pattern complexity. Based on the previously collected pre-trained image-EEG data, we found that these two parameters are strongly correlated with the EEG decoder. Consequently, in the EEG decoder loss function as below, $f$ represents the normalized sum of the amplitude and frequency of the waveform generated by the model, while $f'$ calculates the combined normalized values of the image luminance and pattern complexity. regenerated by the generator. $\lambda$ is a weighting factor set range of 0.01 to 0.1.

We adapted the Pytorch framework for our network training purposes. Optimization was carried out using the Adam optimizer, configuring the first momentum term as $\beta_1 = 0.999$ and setting the initial learning rate to 0.0002. Throughout all experiments, the batch size was maintained at 64. The training process was limited to a maximum of 10,000 epochs. Due to inadequate image generation quality in the initial phases of training, we chose to first save the model parameters after completing 5,000 epochs. Thereafter, parameter saving occurred at every 200-epoch interval. Additionally, at these intervals, images produced by the generator were also generated and saved. The determination of the optimal model parameters was based on the visual quality and effectiveness of the images generated.

## B Image Pre-check Process

The image pre-check process extract a suite of five distinct features: standard deviation of pixel values, edge count determined through the Sobel operator, energy of the high-frequency component from Haar wavelet decomposition, mean frequency from the Fourier transform, and skewness of the pixel intensity histogram. These features encapsulate various aspects of the texture of the image, the information about the edges, the frequency content, and the distribution characteristics. Based on these feature values, it computes the total pairwise Euclidean distances between the feature vectors of the selected images. This metric serves as a quantitative measure of diversity, with the assumption that a higher sum of distances indicates greater dissimilarity among images.

## C   EEG Recording Experimental and Visual Stimulation Protocol Setup

The wearable EEG acquisition system comprised nine channels, including seven for EEG acquisition, one for reference, and one for bias. The microcontroller unit (MCU) served as the control unit, interfacing with the analog-to-digital converter (ADC) via serial input/output and communicating with the workstation using Bluetooth 5.0 protocol. Raw EEG signals were filtered using a 50 Hz band stop filter and sampled at 250 Hz. The amplitude and amplitude of the steady-state visually evoked potential (SSVEP) at the target frequency of 4 Hz, along with the signal-to-noise ratio (SNR), were analyzed using Fast Fourier Transform (FFT). For each stimulation frequency, the EEG data from each trial were segmented into equal epochs starting from the initial timestamp.

Data collection involved six healthy subjects (aged 22-36 years; 2 females and 3 males). The recordings were conducted in a quiet, dimly lit environment, isolated from known sources of electrical interference. Informed consent was obtained from all participants, and the experimental procedures were approved by the Ethics Committee.

Stimulus images consisted of a square that underwent a on-off 100% temporal contrast modulation, displayed in the center of an LCD screen with a resolution of 2560 × 1600 pixels, positioned 60 cm from the subjects' eyes. The stimuli had a visual angle of approximately 6° × 6°. The stimulus was presented at a flicker frequency of 4 Hz (125 ms on and 125 ms off), which was used to detect the response at this specific frequency. Each iteration comprised 30 trials with a 30 ms interval between trials, with up to 8 iterations conducted.

The normalized index and Amplitude of SSVEP were analysis by Python toolbox. The data was presented by mean±s.t.d. The shielding area in the mean response indicates the 95% confidence interval.

## D   Local Field Potentials Recording Setup

Mice were anesthetized with isoflurane (1% at 1–1.5 l min$^{-1}$) and placed on a warming pad (37 °C) to maintain body temperature during surgery. Eye ointment was applied with a cotton swab to keep the mouse's cornea moist. Subsequently, the head skin was cut and the connective tissue was removed to ensure adequate exposure of the skull. A 2-mm diameter craniotomy was performed on the primary visual cortex in the right hemisphere, starting at 2.3 mm lateral and 3.3 mm posterior to the bregma point. The skull was thinned using a cranial drill and carefully soaked in sterile saline solution. The silicon electrode (Neuropixels 2.0) was implanted into the brain at a depth of 2 mm using a micromanipulator (Scientifica).

Visual stimuli were presented on an LCD screen positioned approximately 20 centimeters from the left eye of the mouse, with the content of the stimuli consistent with those used in human subjects. The collected data were subjected to a low-pass filter at 250 Hz and subsequently downsampled to 1 kHz. Local Field Potentials (LFP) recorded within one second post-stimulation were averaged across 50 repetitions.